\documentclass{article}

\usepackage{microtype}
\usepackage{graphicx}
\usepackage{booktabs} 
\usepackage{subcaption}
\usepackage{stfloats}
\usepackage{pgfplots}
\usepackage{hyperref}

\usepackage[accepted]{synsml2023}

\usepackage{amsmath}
\usepackage{amssymb}
\usepackage{mathtools}
\usepackage{amsthm}

\usepackage[capitalize,noabbrev]{cleveref}

\theoremstyle{plain}

\theoremstyle{definition}

\theoremstyle{remark}

\usepackage[textsize=tiny]{todonotes}

\synsmltitlerunning{A Machine Learning Pressure Emulator for Hydrogen Embrittlement}

\begin{document}

\twocolumn[
\synsmltitle{A Machine Learning Pressure Emulator for Hydrogen Embrittlement}




\begin{synsmlauthorlist}
\synsmlauthor{Minh Triet Chau}{yyy}
\synsmlauthor{João Lucas de Sousa Almeida}{sch}
\synsmlauthor{Elie Alhajjar}{comp}
\synsmlauthor{Alberto Costa Nogueira Junior}{sch}
\end{synsmlauthorlist}

\synsmlaffiliation{yyy}{Independent researcher}
\synsmlaffiliation{comp}{RAND Corporation, USA}
\synsmlaffiliation{sch}{IBM Research Brazil}

\synsmlcorrespondingauthor{Minh Triet Chau}{s6michau@uni-bonn.de}

\synsmlkeywords{Machine Learning, CFD, Physics-informed Machine Learning}

\vskip 0.3in
]



\printAffiliationsAndNotice{} 

\begin{abstract}
 A recent alternative for hydrogen transportation as a mixture with natural gas is blending it into natural gas pipelines. However, hydrogen embrittlement of material is a major concern for scientists and gas installation designers to avoid process failures. In this paper, we propose a physics-informed machine learning model to predict the gas pressure on the pipes' inner wall. Despite its high-fidelity results, the current PDE-based simulators are time- and computationally-demanding. Using simulation data, we train an ML model to predict the pressure on the pipelines' inner walls, which is a first step for pipeline system surveillance. We found that the physics-based method outperformed the purely data-driven method and satisfy the physical constraints of the gas flow system.
\end{abstract}

\section{Introduction}
\label{submission}

Climate change requires clean and effective energy storage to replace gasoline, coal, or natural gas (NG). Batteries are a clean carrier but do not have sufficient energy density for sectors such as cement, steel, and long-haul transport \citep{emma__enabling_2021}. For those industries, one option that has received considerable attention is low-carbon hydrogen \citep{osti_1721803}, which can store a large amount of energy and does not release greenhouse pollutants in combustion. However, the inefficiency of green $H_2$ manufacturing process is one of the biggest obstacles to its dissemination 
\citep{osti_1879231}. While finding an environmentally friendly and affordable way to produce $H_2$ is a long-term task, it should not deny us hydrogen's immediate benefit.

One viable strategy is blending $H_2$ with NG (HCNG) to reduce emissions when burning \citep{osti_1219920}. By increasing the volume of $H_2$ from 0\% to 15\%, up to 50\% reduction in $CO_2$ emission was observed \citep{pandey_performance_2022}. Blends with less than 20\% $H_2$ by volume can be transmitted by repurposing existing natural gas pipelines, which are  67\% cheaper than building new ones \citep{peter2020}. However, a major drawback with repurposed pipelines is during daily consumption, gas pressure may reach excessive values that lead to hydrogen diffusion through the most current pipeline wall materials \citep{_european_union_agency_for_the_cooperation_of_energy_regulators_transporting_2021}. \textcolor{black}{Specifically, due to friction incurred on the inner wall caused by the gas flow, atomic $H$ can permeate into its metal lattice, reducing the stress required for cracks to form}. This phenomenon, known as hydrogen embrittlement (HE), causes pipelines to be prone to leaking $H_2$, which can lead to catastrophic events due to $H_2$ ignition in the presence of air, as well as some other complications like decreasing the upper stratospheric ozone mixing ratios \citep{noauthor_atmospheric_nodate}. Such a risk is currently prohibiting HCNG from becoming more popular. In Germany, where it is most widely adopted, HCNG only accounts for 10\% of demand per capita \citep{DOLCI201911394}.

Preventing HE requires monitoring, operational pressure
management, and pipeline maintenance \citep{osti_1646101}. To the best of our knowledge, few works frame pipeline monitoring works from a data driven perspective \citep{SPANDONIDIS2022104890} while the rest rely on signals from sensors and hardware \citep{du_damage_2016, zhu_gas_2017, aba_petroleum_2021}. \textcolor{black}{To apply ML to this problem, there are two steps involved. The first is addressing the prediction task of the gas flow pressure. The next step is to use the predicted pressure as the input to apply a $H_2$ diffusion model through the pipe wall \citep{Fick1855, HAFSI2018210}  to pinpoint in which segment of the pipe the next leakage will be likely to happen. Out of these two issues, we focus on the turbulence gas flow modeling since it is not only a prerequisite but also a more intricate problem than diffusion.} In this work, we propose a supervised machine learning-based model\footnote{Source code is available at \texttt{https://github.com/minhtriet/hydrogen\_emb}.} that predicts future pressure values from previously observed data. Namely, we implement an Operator Inference prototype for pipe surveillance and contrast it with transformer techniques.


The paper is organized as follows. After this short introduction, we explain the dynamical system and the simulation setup for our experiments in Section $2$. In Section $3$, we discuss the results of these simulations and how they compare to other baselines. We conclude the paper with limitations to our work and some future research directions.

\section{Problem}
\label{section:problem_setup}
\begin{figure*}[!tb]
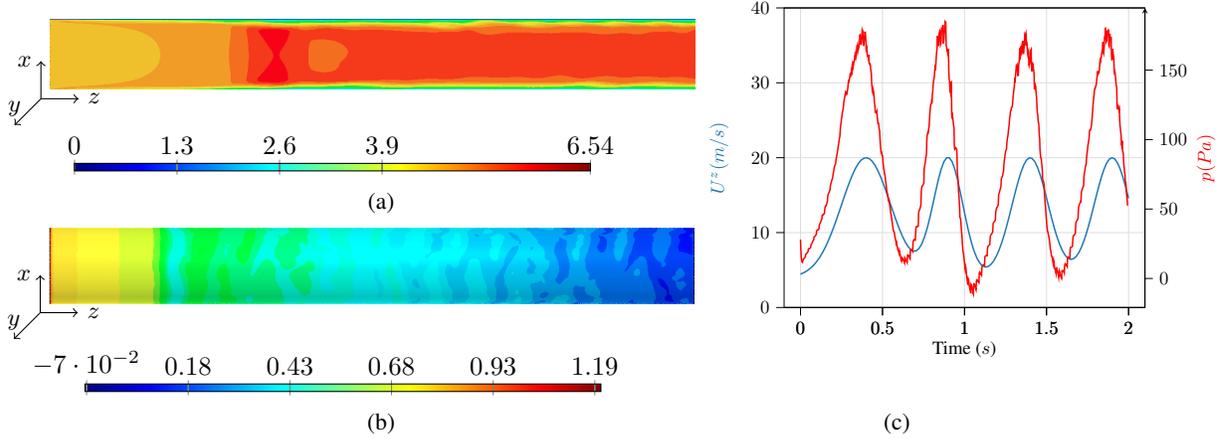

    \centering
    \begin{minipage}[b]{.5\linewidth}
    \begin{subfigure}[b]{\columnwidth}
       \hspace{-9mm} \input{swirl.tikz}
        \caption{}
        \label{subfig:swirl/velocity}
    \end{subfigure}
    \linebreak
    \begin{subfigure}[b]{\columnwidth}
       \hspace{-9mm} \input{swirl_pressure.tikz}
        \caption{}
        \label{subfig:swirl/pressure}
    \end{subfigure}
    \end{minipage}%
    \begin{minipage}[b]{.3\linewidth}
    \begin{subfigure}[b]{\columnwidth}
    \centering
     \input{bc_with_p.tikz} 
     \caption{}
     \label{fig:inlet_gc}
    \end{subfigure}
    \end{minipage}%
    \caption{(\ref{subfig:swirl/velocity}): The velocity field profile $\lVert \textbf{U} \rVert_2$ of gas flow ($m/s$) in a pipeline. (\ref{subfig:swirl/pressure}): The pressure field of gas flow to the inner wall. (\ref{fig:inlet_gc}) The inlet velocity $U^z$ and the mean inner wall pressure $p$ of the whole pipeline through time. The periodic profile for $U^z$ is suitable to imitate  the real-life demand \citep{Su2019}.}
    \label{fig:swirl}

\end{figure*}

\textbf{Physics viewpoint}.
The problem at hand is modeling a turbulent flow. This is because
the Reynolds number of the flow is  $Re=\frac{\rho u D}{\mu} \approx 15,565.58$ Here, $\rho$ is the density, $u$ is the flow speed, $D$ is diameter and $\mu$ is $H_2$ dynamic viscosity.
When the Reynolds number is larger than 2900, the flow is turbulent \citep{Schlichting2016-cw}, which makes the problem chaotic and tricky for ML models to extrapolate (See Figure \ref{fig:swirl}) . For such systems, the most accurate method, as well as time- and computationally-demanding is the Navier-Stokes equation. For a 2D flow, it is defined as \[
    \tau[\frac{\partial U_i}{\partial t}+U_j\frac{\partial U_i}{\partial x_j}] = -\frac{\partial P}{\partial x_i}+\frac{\partial }{\partial x_j}[T_{ij}^{(v)}-{\tau \langle u_iu_j \rangle}]\]
Here, $\delta_{ij}$ is the Kronecker delta, $U_\bullet$ and $u_\bullet$ are the mean and fluctuating velocity, $P$ is the mean static pressure, $T_{ij}^v=2\mu\tilde{s}_{ij}$ is the viscous stresses, $\tilde{s}_{ij}$ $\tilde{s}_{ij}= \frac{1}{2}[\frac{\partial \tilde u_i}{\partial x_j}+\frac{\partial \tilde u_j}{\partial x_i}]$ is the instantaneous strain rate tensor, $\tau\langle u_iu_j\rangle$ is the Reynolds shear stress. 

Assuming a linear  relationship with the mean flow straining field and incompressible flow, $\tau \langle  u_{i} u_{j} \rangle = -2\mu_{t}( S_{ij} - \frac{1}{3} S_{kk} \delta_{ij})$. 
Here $S_{ij}= \frac{1}{2} [ \frac{\partial U_{i}}{\partial x_{j}} + \frac{\partial U_{j}}{\partial x_{i}}]$ is the mean strain rate. To solve it, we use the $k-\omega$ SST model \citep{Menter1993ZONALTE}, a linear eddy-viscosity model. While being highly accurate, solving it requires tremendous time and computation power. 

\textbf{Simulation setup}.
We ran a SimScale simulation with the specifications in Table \ref{tbl:setup} to generate the raw data. Even with only 1000 time steps with fixed $\Delta t=0.002s$, it took 8.4 hours to finish. 

\begin{table} \caption{Simulation setup. The specifications of the pipeline, $U^z$ and $p_\text{outlet}$ are industrial standard from \citep{Mohitpour2003-eg}.}
 \begin{tabular}{@{}ll@{}}
    \toprule
    Parameter & Value\\
    \midrule
    \textit{Pipeline} &\\
    \hspace{3mm}Diameter ($cm$) & 7.62 \\
    \hspace{3mm}Length ($cm$)& 500 \\
    \textit{Boundary condition (BC)} &\\
    \hspace{3mm}Inlet velocity $U^z$ ($m/s$)& 
    Periodic profile in Figure  \ref{fig:inlet_gc}
    \\
    \hspace{3mm}Relative $p_{\text{outlet}}$ $(pa)$ & 0 \\ 
    \hspace{3mm}Wall & No slip \\
    \textit{Simulation control} \\
    \hspace{3mm}$\Delta t$ & 0.002$s$\\
  \bottomrule
    \end{tabular}
    \label{tbl:setup}
\end{table}
\section{Experiments}
\begin{figure*}[!htb]
    \centering\input{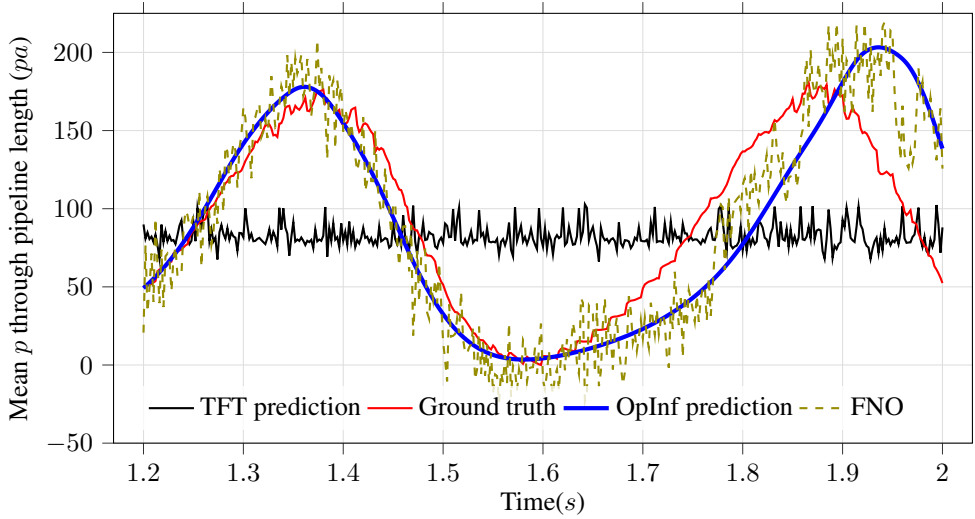}
    \caption{Pressure prediction in the test set.}
\label{fig:less_doesnt_mean_good}
\end{figure*}
\textbf{Set up} \textcolor{black}{For the training data, we select features from the set of (\textbf{U}, $p,k,\omega,nut$) and organize them similarly to the same matrix below, with $m$ the temporal and $n$ the spatial dimension. Each algorithm in our experiment will predict the future $p$, similarly to Figure \ref{subfig:swirl/pressure}. Here, \textbf{U} $\in \mathbb{R}^3$ is the velocity in $x,y,z$ axis, $p$ is the pipeline internal static pressure, $k$ is the rate of dissipation of kinematic turbulence, $\omega$ is the kinematic turbulence energy and $nut$ is the kinematic turbulent viscosity.}
\[
        \begin{pmatrix} 
        k_{1,1} & k_{1,2} & \cdots & k_{1,n} \\[0.5em] 
        p_{1,1} & p_{1,2} & \cdots & p_{1,n} \\[0.5em] 
        \textbf{U}_{1,1} & \textbf{U}_{1,2} & \cdots & \textbf{U}_{1,n} \\[0.5em]  
        \vdots & \vdots & \vdots & \vdots \\[0.5em]
        \textbf{U}_{m,1} & \textbf{U}_{m,2} & \cdots & \textbf{U}_{m,n} \\[0.5em] 
        \end{pmatrix}
\]

The first 50\% of data is for the training set, and the next 10\% is the validation set. After that, we retrain the model with the train and validation sets to predict the test set.
To reduce the computational cost, we first apply PCA to generate a low-rank representation of the dynamic system. 
Then, we apply Transformers, Temporal Fusion Transformers (TFT) \citep{49697}, Operator Inference \citep{ghattas_willcox_2021} and Fourier Neural Operator \citep{https://doi.org/10.48550/arxiv.2010.08895}. The latter two are physics-aware methods in the reduced order space. Each method will predict 10 time steps ahead, then use its own prediction as the input, similarly to a rolling window. At the end of the validation set, each algorithm will immediately produce a $\mathbb{R}^{m_{test} \times n}$ matrix. 

Note that while $U^z$ is available on training, it is absent on testing. \textcolor{black}{Every method has to predict $k$ time-step forward from the end of the training data to the end of testing data without seeing the testing data}   . This is because the $U^z$ is generally not available at the time of running inference to the future.

\textbf{Baseline: Transformer and TFT}. In this approach, we perform a grid search for the hyperparameters, such as the number of encoder and decoder layers, dropout rate, and the number of attention heads. Except for $p$, the other features are set as covariance series. The training of those models is powered by \citep{JMLR:v23:21-1177}. 

\textbf{Fourier Neural Operator}. FNO  aims to learn a neural operator, or neural network, to map from the input to the PDE solution by parameterizing the integral transformation in a Fourier space. The space is chosen to exploit the fact that differentiation is equivalent to multiplication in the Fourier domain.

\textbf{Operator Inference (OpInf)}. \textcolor{black}{Operator Inference assumes a low dimensional manifold from the high fidelity model. It reduces a full order model to a polynomial representation in latent space as
\[\frac{d}{dt}x_t=c+A(x_t)+H(x_t \otimes x_t)+G(x_t \otimes x_t \otimes x_t)+B(u_t)\] 
Here $\otimes$ operator denotes a column-wise Kronecker product. $A, H$ and $G$ represent linear, quadratic, and cubic state matrices respectively.  $B(u_t)$ can be used as the reduced operator of a forcing term in reduced space. $c$ is a constant term. Each of the terms may or may not be used by OpInf, depending on the experimenter's choice. $A$, $H$ and $G$ are learned by solving a regularized least-squares problem. The nonlinear parts of the full-order model are included since many meaningful PDEs have quadratic
nonlinearities in their operators. Combining the linear PCA transformation with the quadratic manifold in latent space is equivalent to applying Galerkin Projection  to the original full-order model using a truncated PCA basis. In this work, we use $A, H, B$ operators and $c$.} 

Besides model selection, we also have to find the optimal number of the bases for its dimension reduction, and the regularizer value for the least square solver since unregularized RMSE minimization may lead to overfitting. For the number of bases, we measure the quality of the reduced order model by computing the preserved energy with respect to the original dynamical system and calculating the RMSE between the original state variables and a PCA reconstruction of those variables. We found out that regardless of the subsets of input ($p, \textbf{U}, k, nut, \omega$), 30 basis capture at least 99.82\% of the cumulative energy of the system. We also perform a PCA on the train and test set with the same number of bases to check if there is a distribution shift between them. The RMSE of the models with their best input set are in Table \ref{tab:exp_transformers}. See Figure \ref{fig:less_doesnt_mean_good} for a visual comparison.
\begin{table*}[!htbp]
    \centering
    \caption{Summary for Transformer, TFT, and OpInf approach. }
    \begin{tabular}{@{}llllll@{}}
    \toprule
    Method & Transformer & TFT & OpInf & FNO \\ 
    \midrule
    Best input feature & $p, U^z$ & $p, U^z$ & $p, U^z$ & $p$\\
    \midrule
     RMSE & 60.5701 & 58.3013 & \textbf{34.7737} & 36.8884 \\
    \bottomrule
    \end{tabular}
    \label{tab:exp_transformers}
\end{table*}

\section{Discussion}
\textcolor{black}{\textbf{The high RMSE of Transformers}. At first, it seems counter-intuitive that a successful model like Transformers behaves like a mean estimator in the test set. In fact, \citep{Zeng2022AreTE} suggests that transformers are ineffective for time-series prediction, even when input data series are measured in years, let alone in a couple of thousands time steps in this work. Indeed, for chaotic systems, the required amount of data is tremendous. For example, \cite{https://doi.org/10.48550/arxiv.2301.10343} uses weather data for ten years to predict reliably for merely one week.} 

Evidently, in Figure \ref{fig:less_doesnt_mean_good}, TFT's high bias leads to failure to capture the system dynamics induced by the BC. On the other hand, we experienced the high variance of OpInf method, where it outputs a sinusoidal prediction but its peaks and valleys are wrong.   
However, decreasing the number of features while keeping the same number of training data took care of the high variance in OpInf. 
It gives the best result as it uses the two most important features $U^z$ and $p$. $p$ is what we are trying to predict, and the flow is strictly dominant in the $z$-axis (See Figure \ref{fig:swirl}). FNO also does a good job.  We observe a jagged prediction of neural network methods. Note the typical prediction of Transformers and TFT, regardless of input features and parameter tuning. For OpInf, a higher regularizer  causes the insensitivity to changes from BC, even though it minimizes test set loss. 

\textcolor{black}{While the simulation is in a short period and small pipeline due to our computational power constraints, it should not hinder its generalization to a bigger pipeline system. It is because the length of the pipe only indicates how pressure drops along the lines, and the pressure equation should follow the same physics equations. For the time-bound question, OpInf has demonstrated its accuracy in complex systems with long prediction window \citep{https://doi.org/10.48550/arxiv.2203.06361} or a high number of snapshots \citep{Swischuk_2020, DBLP:journals/corr/abs-2008-02862}. } 

\textcolor{black}{\textbf{The importance of speeding up with ML}. Taking a step back, the HE problem is a control problem, as it concerns what to do when a breach is happening. Therefore, to predict when and where the next leakage might occur, it is best to create a handful of scenarios with different $U,p$ profiles in different pipeline systems. However, turbulence modeling is expensive, even in a pipeline, as noted in Section \ref{section:problem_setup}. In comparison, the inference time in ML methods is less than a second compared to 8.4 hours in the numerical simulation, making the incorporation of such ML techniques crucial in future investigations.}

\textcolor{black}{\textbf{The degradation of the prediction performance.} In Figure \ref{fig:less_doesnt_mean_good}, the RMSE is increasing around $t=1.9s$, which hints at a limitation of the prediction window for physics-aware algorithms. With the current amount of data in our experiments, we cannot quantify the relationship between the size of the dataset and the prediction window length, as well as the confidence interval of the training regime. Both of these pillars will be a part of a follow up paper. Moreover, quantifying how the error in the pressure prediction affects the HE depends on the $H_2$ diffusion models, which we already discussed in Section $1$.}

\section{Conclusion}
In this work, we propose an ML model for emulating HE, which is one of the biggest obstacles against HCNG and $H_2$ public usage. From an environmental perspective, wider adoption of HCNG can reduce the amount of $CO_2$ exhaustion. Moreover, preventing $H_2$ leakage also improves the safety of the transport system and the atmosphere quality of the surrounding area. Compared to numerical simulations, our ML model takes less than a second to produce an accurate inference of the pressure in a pipeline. It is the first step to bring HCNG one step closer to mass adoption today and pure $H_2$ delivery infrastructure in the future.

\bibliographystyle{synsml2023}
\bibliography{example_paper}


\end{document}